\documentclass[letterpaper]{article} 
\usepackage{aaai24}  
\usepackage{times}  
\usepackage{helvet}  
\usepackage{courier}  
\usepackage[hyphens]{url}  
\usepackage{graphicx} 
\usepackage{natbib}  
\usepackage{caption} 
\usepackage{algorithm}
\usepackage{algorithmic}

\usepackage{newfloat}
\usepackage{listings}
\DeclareCaptionStyle{ruled}{labelfont=normalfont,labelsep=colon,strut=off} 
\floatstyle{ruled}
\newfloat{listing}{tb}{lst}{}
\floatname{listing}{Listing}
\usepackage{amsmath}
\usepackage{amssymb}
\usepackage[table,xcdraw]{xcolor}
\usepackage{pifont}
\usepackage[caption=false]{subfig}
\usepackage{fancyhdr}

\newcommand{\cmark}{\ding{51}}%
\newcommand{\xmark}{\ding{55}}%
\DeclareMathOperator*{\argmax}{arg\,max}

\title{Uncertainty Propagation through Trained \\ Deep Neural Networks Using Factor Graphs}
\author{
    Angel Daruna,
    Yunye Gong,
    Abhinav Rajvanshi,
    Han-Pang Chiu,
    Yi Yao
}
\affiliations{
    \{angel.daruna,yunye.gong,abhinav.rajvanshi,han-pang.chiu,yi.yao\}@sri.com
    
    Center for Vision Technologies\\SRI International\\Princeton, NJ
    
}

\begin{document}

\maketitle

\begin{abstract}
Predictive uncertainty estimation remains a challenging problem precluding the use of deep neural networks as subsystems within safety-critical applications. Aleatoric uncertainty is a component of predictive uncertainty that cannot be reduced through model improvements. Uncertainty propagation seeks to estimate aleatoric uncertainty by propagating input uncertainties to network predictions. Existing uncertainty propagation techniques use one-way information flows, propagating uncertainties layer-by-layer or across the entire neural network while relying either on sampling or analytical techniques for propagation. Motivated by the complex information flows within deep neural networks (e.g. skip connections), we developed and evaluated a novel approach by posing uncertainty propagation as a non-linear optimization problem using factor graphs. We observed statistically significant improvements in performance over prior work when using factor graphs across most of our experiments that included three datasets and two neural network architectures. Our implementation balances the benefits of sampling and analytical propagation techniques, which we believe, is a key factor in achieving performance improvements.
\end{abstract}

\section{Introduction}

Aleatoric uncertainty estimation of deep neural network predictions is a challenging problem precluding their use within saftey critical applications. Neural networks present a new method for processing physical sensor data, improving over traditional methods in many domains, e.g. inertial odometry \cite{liu2020tlio}. Despite such advancements, extracting predictive uncertainty estimates from trained neural networks remains a challenge. As a result, incorporating neural networks within safety-critical applications that combine many predictions and their uncertainties remains an open question. Predictive uncertainty is typically modeled as two separate uncertainties, epistemic and aleatoric uncertainty. Aleatoric uncertainty stems from environmental variations and sensor noise; hence, aleatoric uncertainty cannot be reduced through model improvements \cite{gawlikowski2023survey}. 

In this paper, we examine a new technique to estimate the aleatoric uncertainty of a trained deep neural network, using uncertainty propagation. Some prior methods for aleatoric uncertainty estimation augment training procedures, which can improve the prediction performance of a trainable neural network. Examples include Bayesian Deep Learning \cite{kendall2017uncertainties}, Ensemble Distribution Distillation \cite{malinin2019ensemble}, and Evidential Neural Networks \cite{sensoy2018evidential,amini2020deep}. Other prior work has sought to assess the aleatoric uncertainty at inference time for a \textit{not-editable trained} neural network, known as input uncertainty propagation \cite{titensky2018uncertainty,monchot2023input}. In this way, uncertainty propagation techniques can be applied to a trained neural network without augmentation. Existing uncertainty propagation techniques use one-way information flows, either propagating uncertainties layer-by-layer or across the network while relying either on sampling or analytical techniques for propagation.
\begin{figure}[t]
    \centering
    \includegraphics[width=0.8\columnwidth]{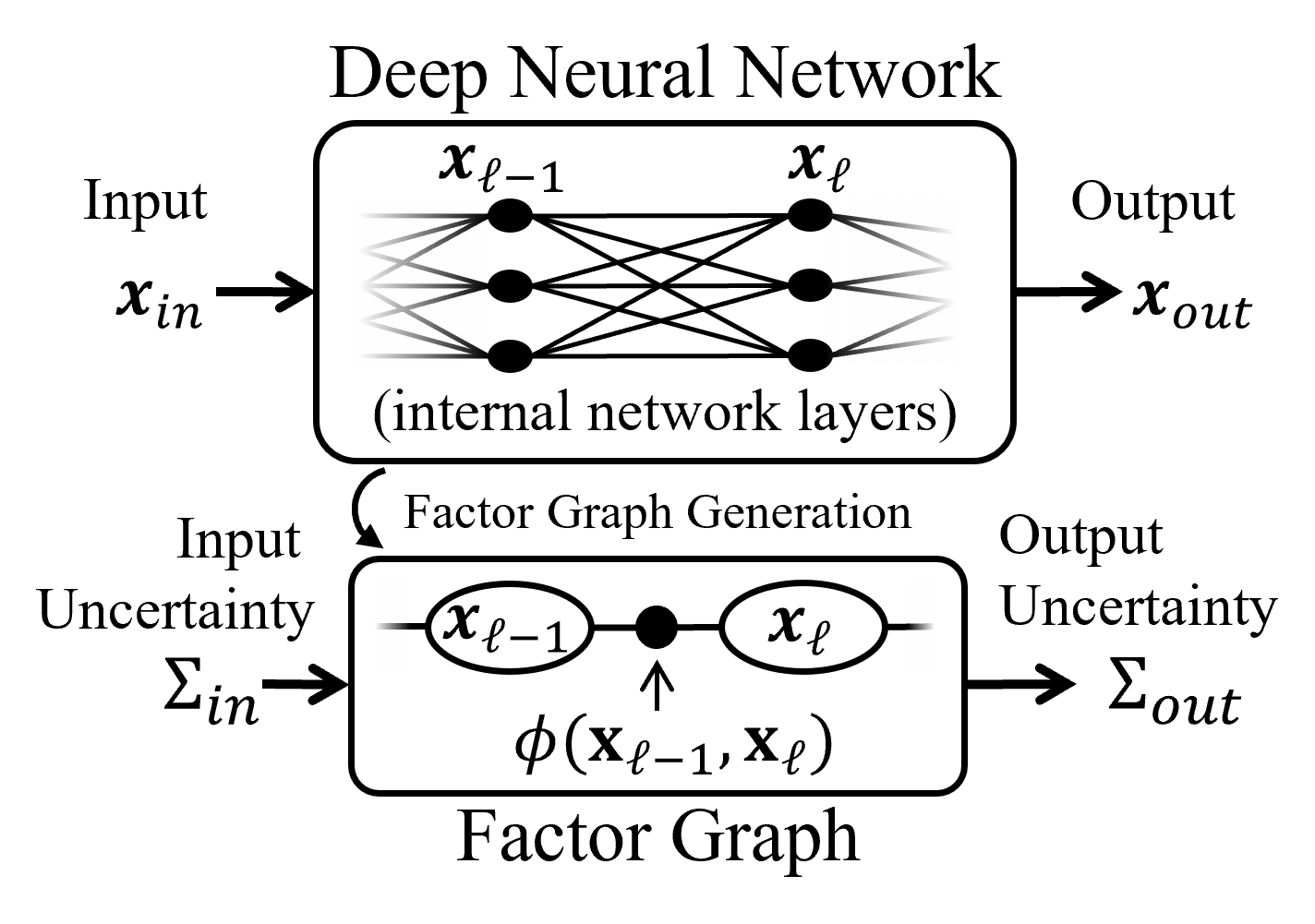} 
    \caption{Our uncertainty propagation approach models deep neural networks using factor graphs to estimate predictive uncertainty caused by data uncertainty.}
    \label{figs:overview}
\end{figure}

We develop and validate a factor graph formulation for modeling deep neural network uncertainty propagation as a non-linear optimization problem by treating network layers as discrete time steps, layer values as variable nodes, and connections among layers as factors (Figure~\ref{figs:overview}). Factor graphs \cite{kschischang2001factor} are a probabilistic Bayesian graphical model that factorize a probability density over a set of factors involving variable nodes. While factor graphs have been widely used for large-scale real-time state estimation, to the best of our knowledge, we are the first to leverage factor graphs for neural network uncertainty estimation. Factor graphs can be used to address several limitations of prior uncertainty propagation works. These limitations include limited information flow when estimating uncertainty as well as balancing the benefits of sampling from the input uncertainty distribution \cite{abdelaziz2015uncertainty} and analytically propagating network uncertainties \cite{titensky2018uncertainty}.

We evaluated our uncertainty propagation approach against three baselines for classification and regression problems. Our experiments span three datasets (MNIST \cite{lecun-mnisthandwrittendigit-2010}, CIFAR-10 \cite{krizhevsky2009learning}, and M2DGR \cite{yin2021m2dgr}) and two neural network architectures, (Multi-Layer Perceptron and ResNet18 \cite{he2016deep}). In each experiment, we trained a deep neural network for a task on the correspondent dataset. Then we evaluated how well each uncertainty estimation method could propagate simulated or actual input uncertainties. When ground truth was unavailable, input uncertainties were simulated in four different noise settings by varying the amount of white noise and local correlation. We demonstrate the differences in the uncertainty estimations of each technique both quantitatively using the 2-Wasserstein distance calculated over many samples and qualitatively observing the output uncertainties. Across all experiments, we observed improved performance from using factor graphs for propagating input uncertainty when compared to the baselines. In all but a single experimental case, our statistical analyses of results using a non-parametric one-way repeated measures analysis of variance indicated improvements were statistically significant (p-value $=0.001$). Our implementation balances the benefits of sampling and analytical propagation techniques, which we believe, is a key factor in achieving performance improvements.

In summary, our work contributes:
\begin{enumerate}
    \item To the best of our knowledge, the first factor graph formulation for trained neural network uncertainty estimation, specifically, input uncertainty propagation.
    \item An implementation that balances between sampling and analytical uncertainty propagation techniques.
    \item An evaluation demonstrating that our approach outperforms prior methods according to several experiments across multiple datasets and network architectures.
\end{enumerate}

\section{Related Works \& Background}

Our work is motivated by prior research in two areas: Uncertainty Propagation and Factor Graphs.

\subsection{Input Uncertainty Propagation}

Uncertainty Propagation in deep learning seeks to estimate the predictive uncertainty induced by aleatoric uncertainty. Several prior works propagate input uncertainties by modeling the input and output distributions as Gaussian. The first two moments have been propagated both layer-wise \cite{astudillo2011propagation} and across entire neural networks \cite{abdelaziz2015uncertainty} using the unscented transform. Lightweight Probabilistic Networks were used in \cite{gast2018lightweight} to propagate the first two moments analytically while ignoring correlations of weight dimensions within layers (i.e. diagonal covariance matrices). The Extend Kalman Fitering formulation of \cite{titensky2018uncertainty} enabled the analytic propagation of the first two moments layer-wise while modeling correlations of weight dimensions within layers (i.e. full covariance matrix). The work of \cite{monchot2023input} propagated input uncertainties by instead modeling the input and output distributions as Gaussian mixtures and propagating uncertainties layer-by-layer or across the deep neural network. Motivated by the complex information flows within deep neural networks that are not modeled by prior methods (e.g. skip connections), we developed and evaluated a novel method by posing uncertainty propagation as a non-linear optimization problem using factor graphs.

\subsection{State Estimation using Factor Graphs}

Factor Graphs (FG) are probabilistic Bayesian graphical models that naturally encode the factored nature of probability densities over multiple variables and their interactions (i.e. factors) \cite{kschischang2001factor}. Therefore, a factor graph can be used to evaluate the probability density of a state, find a local maximum of a posterior distribution, and sample states from a probability density \cite{dellaert2017factor}. Factor graphs are used in many prior works as a powerful tool for efficiently solving estimation problems. In \cite{indelman2012factor} factor graphs are used along with the incremental smooth and mapping algorithm to formulate an inertial navigation pipeline. In \cite{whelan2015real} factor graphs are leveraged within a dense simultaneous localization and mapping solution to produce dense 3D maps that can enable more complex environment interactions, such as manipulation. The approach of \cite{johannsson2010imaging} relies on a factor graph framework to aid inertial odometry with sonar measurements for underwater navigation.

To the best of our knowledge, we are the first to use factor graphs to model deep neural network uncertainty propagation. The next section introduces our formulation.

\section{Approach}

We use factor graphs to model deep neural network uncertainty propagation as a non-linear optimization problem. We first instantiate a corresponding factor graph for a trained deep neural network that is the target for uncertainty propagation. The factor graph is formulated by treating network layers as discrete time steps and their values as variable nodes. We model connections among layers as different factors across variable nodes. The Jacobian matrix and the noise matrix for each factor comes from the weights and biases across correspondent layers of the trained deep neural network. The output covariance of the deep neural network can be accurately estimated by propagating input uncertainty through the network using the factor graph variable nodes and factors. Below we provide more details of this algorithmic overview.

\begin{figure}[t]
    \centering
    \includegraphics[width=0.8\columnwidth]{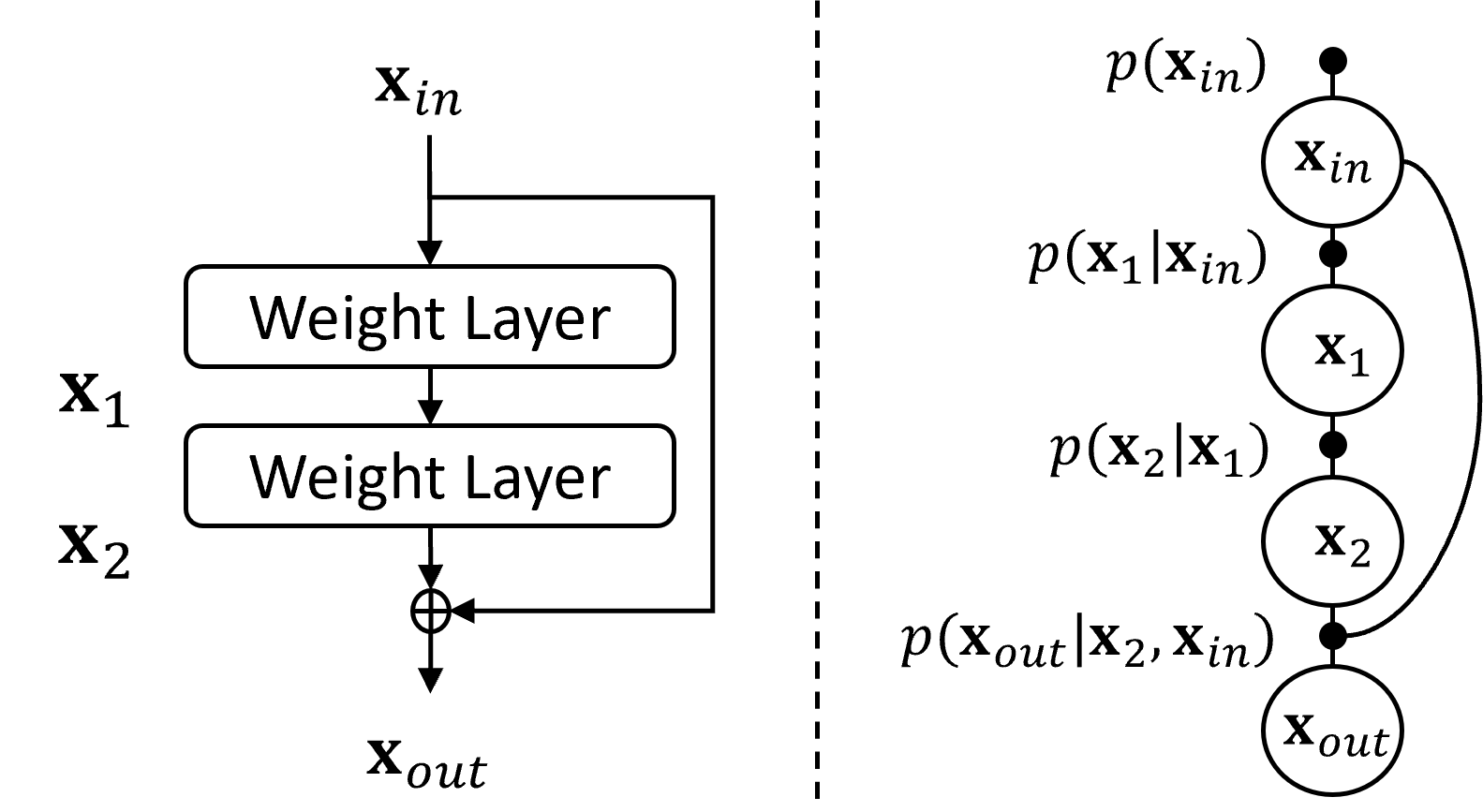}
    \caption{Example of factor graph (right) that can be used to propagate input uncertainty for a residual-block of a trained ResNet (left).}
    \label{figs:fg-block}
\end{figure}

\subsection{Factor Graph Formulation}

We use a factor graph formulation to model deep neural network uncertainty propagation, by treating the network layers as discrete time steps and their values as variable nodes. Formally a factor graph is a bipartite graph $F = (\mathcal{U},\mathcal{V},\mathcal{E})$ with two types of nodes: factors $\phi_{i} \in \mathcal{U}$ and variables $x_j \in \mathcal{V}$ \cite{kumar2012plug}. Edges $e_{ij} \in \mathcal{E}$ encode independence relationships and are \textit{always} between factor nodes and variables nodes. A factor graph $F$ defines a factorization of a global function, $\phi(X)$ in Equation~\ref{eqs:fg}, where the factors are functions of only adjacent variables, $X_i$ in Equation~\ref{eqs:fg}, connected via edge $e_{ij}$.
\begin{equation} \label{eqs:fg}
    \phi(X)=\prod_i \phi_i(X_i)
\end{equation}

We instantiate a factor graph corresponding to the structure of a trained deep neural network that is the target for uncertainty propagation. In Figure~\ref{figs:fg-block}, we provide a small example showing a factor graph that would correspond to a residual-block of a ResNet \cite{he2016deep}. Features of each layer $j$ in the deep neural network are represented as variable node vectors $x_j$. Connections among network layers are formulated as different factors $\phi_i(X_i)$ among adjacent variable nodes. In other words, we formulate a factor graph $F = (\mathcal{U},\mathcal{V},\mathcal{E})$ which models the input, output, and intermediate features in a deep neural network as variable nodes $x_j \in \mathcal{V}$ and defining different factor nodes $\phi_{i} \in \mathcal{U}$ based on layer connections (i.e. edges $e_{ij}$ within the factor graph).

\begin{figure}[b!]
    \centering
    \includegraphics[width=0.9\columnwidth]{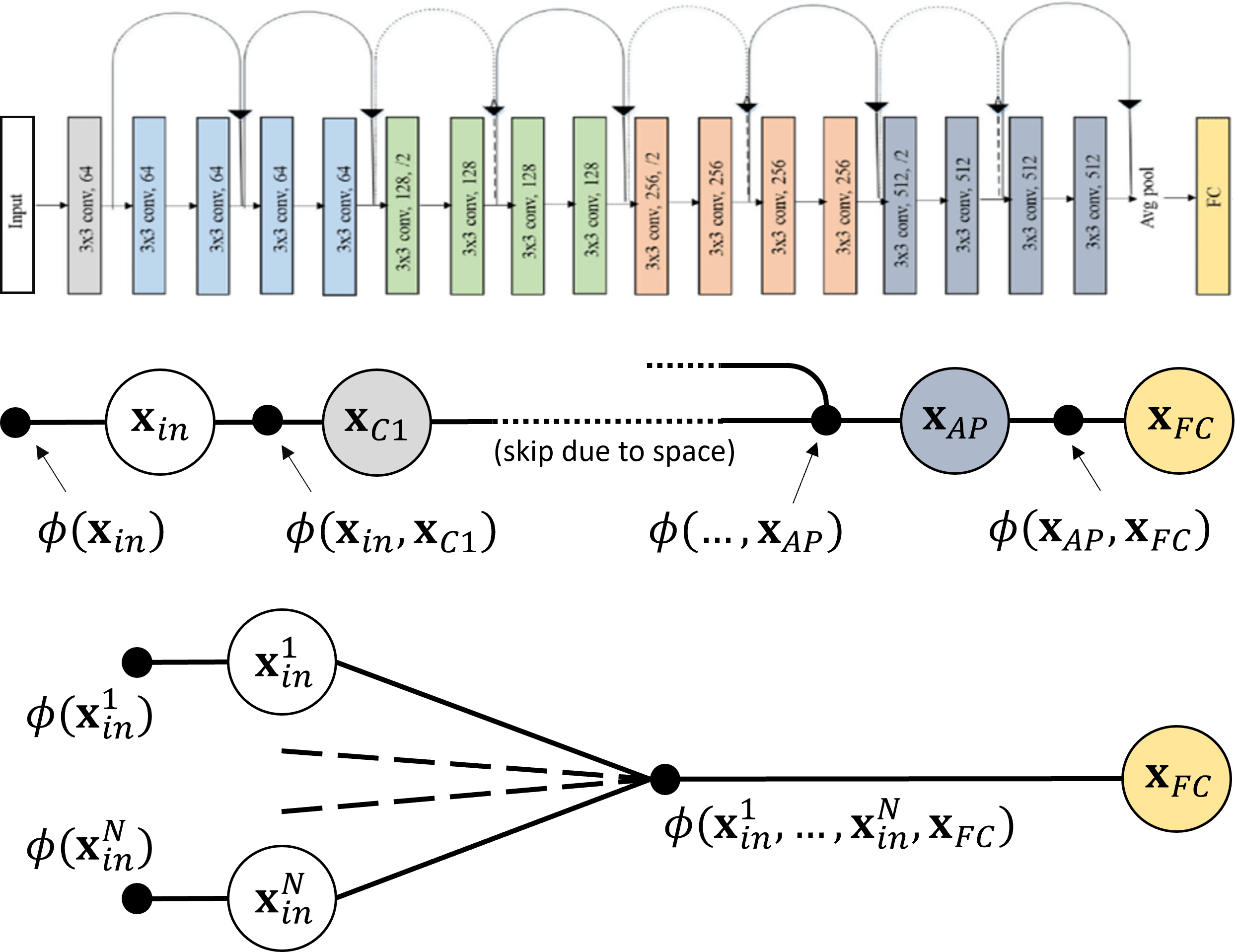} 
    \caption{Example of factor graphs (below) that can be used to propagate input uncertainty for a trained deep neural network, ResNet18 (above).}
    \label{figs:correspondence}
\end{figure}

As a concrete example, we discuss factor graph formulations to propagate the input uncertainty of ResNet18 \cite{he2016deep}, a commonly used deep neural network architecture. Many different factor graphs can model the same deep neural network using a different factorization of the probability density. Our initial factor graph formulation for ResNet18 included nodes for every layer of ResNet18 with factors for all layer connections, as introduced in the previous paragraph with Figure~\ref{figs:fg-block}. The first factor graph in Figure~\ref{figs:correspondence} shows part of the complete graph instantiated for ResNet18 that has been truncated due to space.

After initial experiments, we settled on a reduced formulation shown in the last factor graph of Figure~\ref{figs:correspondence}. We observed that modeling all high dimensional layers of ResNet18 simultaneously was computationally prohibitive\footnote{Existing factor graph modeling libraries are CPU only}. Therefore, we opted for a factor graph that uses two sets of factor graph nodes, one set represents the state of the input to ResNet18, the other represents the state of the target layer for uncertainty propagation (i.e. output of ResNet18’s fully connected layer, FC in Figure~\ref{figs:correspondence}). We use only two types of factors modeled as Gaussian densities, prior factors and between factors, defined in Equations~\ref{eqs:prior_factor} and \ref{eqs:between_factor}, respectively. Prior factors estimate the input uncertainty prior. The between factor estimates the uncertainty between the input and output variable nodes being modeled. In Equations~\ref{eqs:prior_factor} and \ref{eqs:between_factor}, to save space, we have defined $\eta(\Sigma) = \frac{1}{\sqrt{|2\pi\Sigma|}}$ as a normalizer and $||\theta-\mu||_\Sigma^2\triangleq(\theta-\mu)^T \Sigma^{-1} (\theta-\mu)$ as the squared Mahalanobis distance with covariance matrix $\Sigma$.
\begin{equation} \label{eqs:prior_factor}
    \phi(x_{in})=\eta(\Sigma_{in})\exp\Big(\textrm{-}\frac{1}{2}||x_{in}-\mu_{in}||^2_{\Sigma_{in}}\Big)
\end{equation}
\begin{equation} \label{eqs:between_factor}
    \phi(x_{in},x_{\textrm{FC}})=\eta(\Sigma_{\textrm{FC}})\exp\Big(\textrm{-}\frac{1}{2}||r_{18}(x_{in})-x_{\textrm{FC}}||^2_{\Sigma_{\textrm{FC}}}\Big)
\end{equation}

Empirically, we observed improved uncertainty propagation performance by using more than a single input node in the factor graph. While Equation~\ref{eqs:between_factor} is written for a single input node given available space, the binary between factor is extended to an n-ary between factor by taking the sum of differences between input nodes and the output node. 

Our uncertainty propagation approach can be applied to any layer of a trained neural network without affecting architecture complexities or training procedures. Following a similar procedure described above, we can instantiate many different reduced factor graphs to propagate input uncertainties to any intermediate layers of ResNet18, e.g. 2D adaptive average pooling layer labeled $AP$ in Figure~\ref{figs:correspondence}. The neural network architecture complexities and training procedures are unaffected by our uncertainty propagation technique because the factor graph is external to the trained neural network. Network augmentation or retraining is not necessary.

\subsection{Uncertainty Propagation}

With the factor graph structure defined, uncertainty propagation can begin after initializing the values and uncertainties for all variable nodes and Jacobians for all factors. While factor graphs are not limited by the choice of distribution to model variable nodes \cite{kaess2012isam2}, we assume Gaussian densities for variable nodes to leverage existing factor graph libraries \cite{gtsam}. The initial value of each variable node within the factor graph is initialized to the corresponding value within the deep neural network or a sampled input value, according to the variable node set. We initialize the uncertainties for variable nodes (i.e. covariances) using either the given input uncertainty (e.g., input variable nodes) or using an identity matrix with diagonal terms scaled close to zero for unknown uncertainties (e.g., output variable nodes). The Jacobian matrix for each factor comes from the weights and biases for correspondent layers of the trained deep neural network.

After defining the factor graph structure and initializing variable nodes, uncertainty propagation is used to estimate the predictive uncertainty caused by the input noise. Inference with covariance propagation over a factor graph corresponds to finding the maximum a posteriori (MAP) estimate of the variable nodes. Each factor encodes a likelihood that should be maximized by adjusting the estimates of the involved variable nodes. The optimal estimate is the one that maximizes the likelihood of the entire graph, $\phi(X)$:
\begin{equation} \label{eqs:MAP}
    \hat{X}=\argmax_{x}\Big(\prod_i \phi_i(X_i)\Big)
\end{equation}

Maximizing the factored representation in Equation~\ref{eqs:MAP} is equivalent to minimizing the negative log likelihood. In the Gaussian case, this leads to a nonlinear least-squares problem. To solve the full nonlinear optimization problem underlying the factor graph framework, the incremental smoothing and mapping inference engine provides an efficient solution using the Bayes tree structure, which can be obtained from the factor graph by variable elimination \cite{kaess2012isam2}.

\section{Experiments}

We evaluated our uncertainty propagation approach against three baselines across classification and regression problems. Our experiments span three datasets (MNIST \cite{lecun-mnisthandwrittendigit-2010}, CIFAR-10 \cite{krizhevsky2009learning}, and M2DGR \cite{yin2021m2dgr}) and two neural network architectures (Multi-Layer Perceptron and ResNet18). In each experiment, we trained a deep neural network for a task on the correspondent dataset, then evaluated how well each approach could propagate input uncertainties. Across all experiments, we observed improved performance from using our approach for propagating input uncertainty when compared to the baselines. In all but a single experimental case, our statistical analyses of results indicated improvements were statistically significant ($p = 0.001$). Below we provide more details of our experimental procedure and results for each experiment.

\subsection{Experimental Procedure}

We evaluated uncertainty propagation performance by comparing the output uncertainties estimated in each method to a ground truth output uncertainty reference. We tested a variety of baselines from related works in addition to our approach. The lightweight probabilistic network baseline (\textbf{LPN} in figures and tables)
trains a lightweight probabilistic deep neural network to predict the variance for each network output \cite{gast2018lightweight}. However, variance alone ignores the off diagonal terms of network output uncertainties. The unscented transform (\textbf{UT} in figures and tables) and Extended Kalman Fitler (\textbf{EKF} in figures and tables) baselines use the unscented transform and extended Kalman fitler, repsectively, to estiamte the full covariance of network output uncertainties \cite{abdelaziz2015uncertainty,titensky2018uncertainty}. We used Monte Carlo sampling as a reference for the network's true output uncertainty by computing the sample covariance of network outputs generated from sampling the input uncertainty distribution. The 2-Wasserstein distance is a metric to describe the distance between two distributions, in our case, Gaussians. We compare each method to the Monte Carlo sampling reference by computing the 2-Wasserstein distance between the estimated covariance matrices of each.

\begin{figure}[t]
    \centering
    \includegraphics[width=0.98\columnwidth]{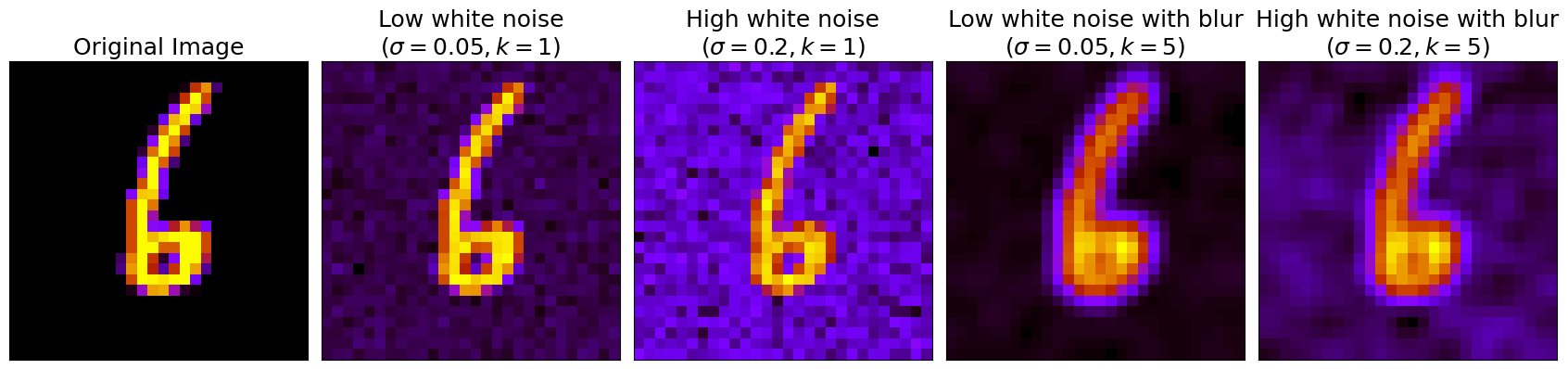}
    \caption{Examples of experimental cases using white and blur noises applied to a single MNIST image.}
    \label{figs:noise}
\end{figure}

In our image classification experiments, controlled input uncertainties were simulated to evaluate each uncertainty propagation technique because ground truth data was unavailable (i.e. true pixel value). We corrupted input images by assuming each measured pixel value was drawn from a Gaussian distribution centered at the true pixel value with some standard deviation $\sigma$, and blurring the image with a varying kernel size $k$. The combination of these noise types effectively results in corrupting the image with additive Gaussian noise that may be locally correlated when blurring is included (i.e. diagonal or semi-diagonal covariance matrix). For each experiment we evaluated four experimental cases from the combinations of low ($\sigma=0.05$) or high ($\sigma=0.2$) white noise and whether blurring did ($k=5$) or did not occur ($k=1$). An example of each experimental setting applied to an image from MNIST is shown in Figure~\ref{figs:noise}. Below we provide details specific to each experiment followed by a summary of results.

\subsection{MNIST Classification with Multi-Layer Perceptron}

As a first experiment, we propagated input uncertainties for a four layer Multi-Layer Perceptron (MLP) with ReLU activations trained to classify MNIST digits. We trained the MLP to an accuracy of 98\% using an 80/20 split for training and testing with uncorrupted images. We then evaluated each uncertainty propagation approach against the Monte Carlo sampling reference using 2-Wasserstein distances for 500 example images from our MNIST test set with each of the controlled noise settings previously discussed. We repeated this evaluation for both the MLP output layer and the third layer of the MLP to demonstrate that our approach can propagate input uncertainties to any layer within the target network by altering the factor graph. To test for statistical significance in each experiment setting, we performed a non-parametric one-way repeated measures analysis of variance (Friedman's test) followed by a post-hoc analysis to identify which experimental groups differ (Nemenyi's test). Friedman's and Nemenyi's tests were used because the 2-Wasserstein distances were not normally distributed, according to Shapiro-Wilk tests. The results are presented in the next section.

\begin{figure}[t]
    \centering
    \includegraphics[width=0.95\columnwidth]{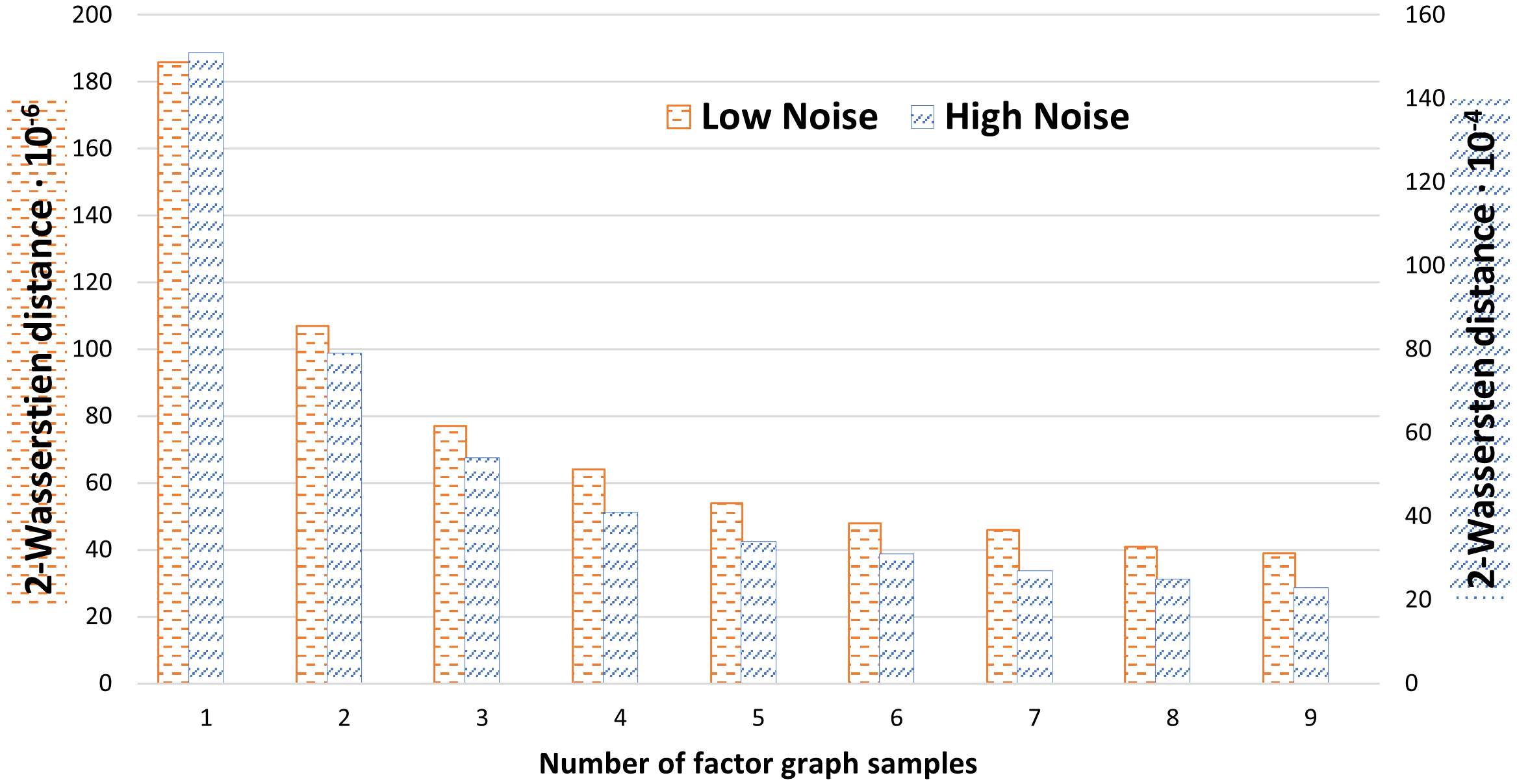}
    \caption{Ablation study showing diminishing returns of using a factor graph with more than 4-5 samples.}
    \label{figs:ablation}
\end{figure}

\begin{table}[t]
    \begin{tabular}{c|c|r|r|r|l}
        \hline
        \textbf{\begin{tabular}[c]{@{}c@{}} \small Noise \\ \small Level\end{tabular}} & \small \textbf{Blur} & \multicolumn{1}{c|}{\small \textbf{LPN}} & \multicolumn{1}{c|}{\small \textbf{UT}} & \multicolumn{1}{c|}{\small \textbf{EKF}} & \textbf{\begin{tabular}[c]{@{}c@{}} \small FG \\ \small (ours)\end{tabular}} \\
        \hline
        \multicolumn{6}{c}{MLP Output Layer} \\
        \hline
        \small Low & \xmark & \small 5.49654 & \small 0.00038 & \small 0.00009 & \small 0.00006 \\
        \small Low & \cmark & \small 5.41811 & \small 0.00439 & \small 0.00425 & \small 0.00004 \\
        \small High & \xmark & \small 8.74004 & \small 0.02857 & \small 0.00499 & \small 0.00412 \\
        \small High & \cmark & \small 6.90933 & \small 0.07487 & \small 0.06577 & \small 0.00209 \\
        \hline
        \multicolumn{6}{c}{MLP Third Layer} \\
        \hline
        \small Low & \xmark & \small N/A & \small 0.00131 & \small 0.00025 & \small 0.00018 \\
        \small Low & \cmark & \small N/A & \small 0.00591 & \small 0.00552 & \small 0.00009 \\
        \small High & \xmark & \small N/A & \small 0.09391 & \small 0.01504 & \small 0.01144 \\
        \small High & \cmark & \small N/A & \small 0.10873 & \small 0.08257 & \small 0.00467 \\
        \hline
    \end{tabular}
    \caption{MNIST Uncertainty Propagation Performance for MLP 3rd layer and output layer measured using 2-Wasserstien Distance (lower is better)}
    \label{tbl:mnist}
\end{table}

Before evaluating uncertainty propagation, we performed an ablation study using the validation set to select an appropriate number of inputs nodes for our factor graph approach. We propagated the input uncertainties using our factor graph approach for 500 example images from our MNIST validation set in both the low and high white noise setting without blurring. The factor graph estimated output uncertainties were compared to the Monte Carlo sampling reference using 2-Wasserstein distance. We repeated this procedure for 1 to 9 input variable nodes within the factor graph. We observed that using more than 4-5 input variable nodes yields diminishing returns, shown in Figure~\ref{figs:ablation}. Specifically, the rate of performance improvement is less than 20\% when using 4 or more samples and every additional sample increases the uncertainty propagation time. We use 4 factor graph samples in all later experiments for this reason.

\begin{table}[t]
    \centering
    \begin{tabular}{c|c|r|r|l}
        \hline
        \textbf{\begin{tabular}[c]{@{}c@{}} \small Noise \\ \small Level\end{tabular}} & \small \textbf{Blur} & \multicolumn{1}{c|}{\small \textbf{UT}} & \multicolumn{1}{c|}{\small \textbf{EKF}} & \textbf{\begin{tabular}[c]{@{}c@{}} \small FG \\ \small (ours)\end{tabular}} \\
        \hline
        \small Low & \xmark & \small 10.16012 & \small 0.00390 & \small 0.00211 \\
        \small Low & \cmark & \small 10.03735 & \small 0.03680 & \small 0.00016 \\
        \small High & \xmark & \small 45.06825 & \small 1.00291 & \small 0.41753 \\
        \small High & \cmark & \small 10.60831 & \small 0.50112 & \small 0.02940 \\
        \hline
    \end{tabular}
    \caption{CIFAR-10 Uncertainty Propagation Performance for ResNet18 measured using 2-Wasserstien Distance (lower is better)}
    \label{tbl:cifar}
\end{table}

\subsubsection{Experimental Results}

Uncertainty propagation using factor graphs provided statistically significant (p-value $=0.001$) improvements in performance against all baselines in our MNIST experiments, as summarized by Table~\ref{tbl:mnist}. Lightweight probabilistic neural networks (LPN baseline) show the worst performance across experimental cases because the approach trains a network only to provide variances of the network outputs, which ignores off-diagonal terms within covariance matrices of output network uncertainties. Additionally, LPN was not evaluated in the MLP third layer experiment as it only provides uncertainties for network outputs. The UT baseline relies on sampling sigma points while the EKF baseline relies on analytical formulas to propagate the input uncertainty; both underperformed against our factor graph implementation. We believe our factor graph formulation is able to outperform the baselines because it incorporates both sampling and analytical techniques to propagate uncertainties, in addition to non-linear optimization over the entire graph.

\begin{figure*}[t]
    \centering
    \includegraphics[clip,width=1.05\columnwidth]{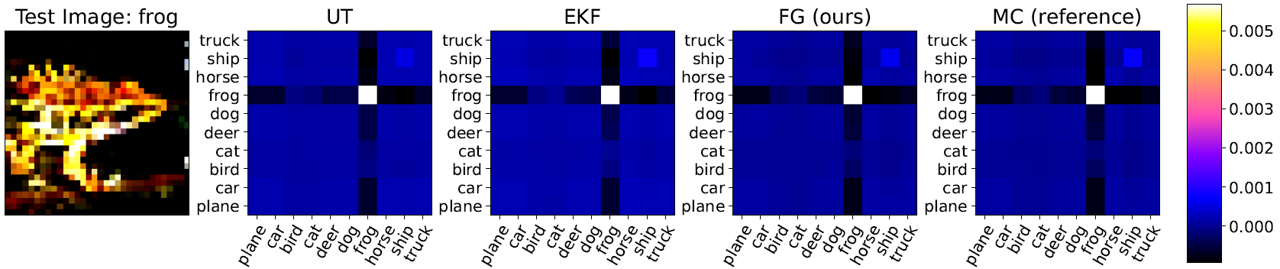}\hfill
    \includegraphics[clip,width=1.05\columnwidth]{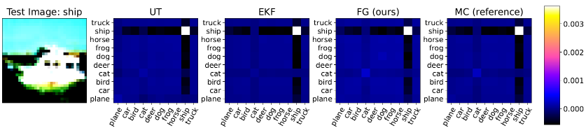}
    
    \includegraphics[clip,width=1.05\columnwidth]{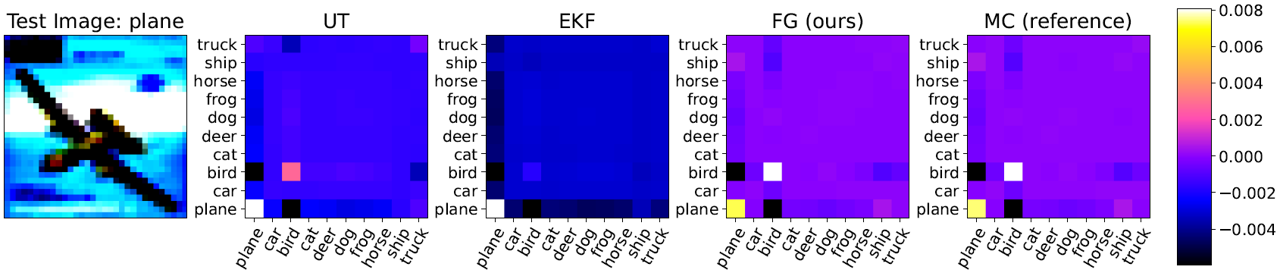}\hfill
    \includegraphics[clip,width=1.05\columnwidth]{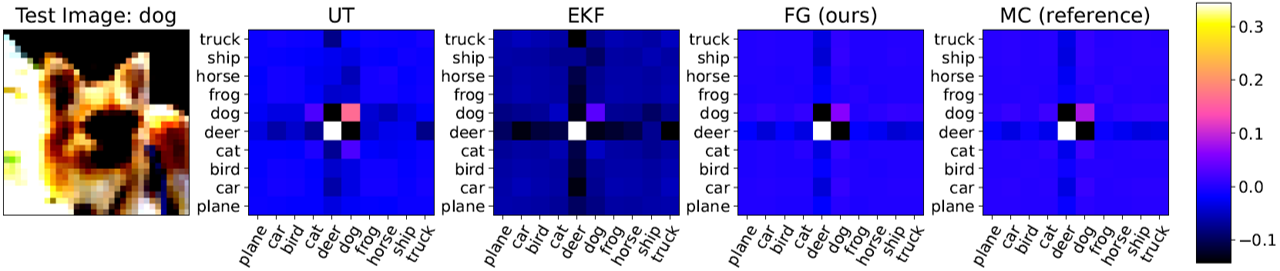}
    
    \includegraphics[clip,width=1.05\columnwidth]{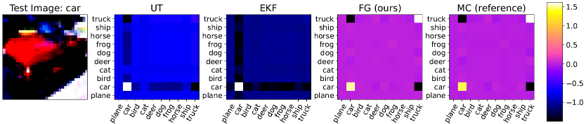}\hfill
    \includegraphics[clip,width=1.05\columnwidth]{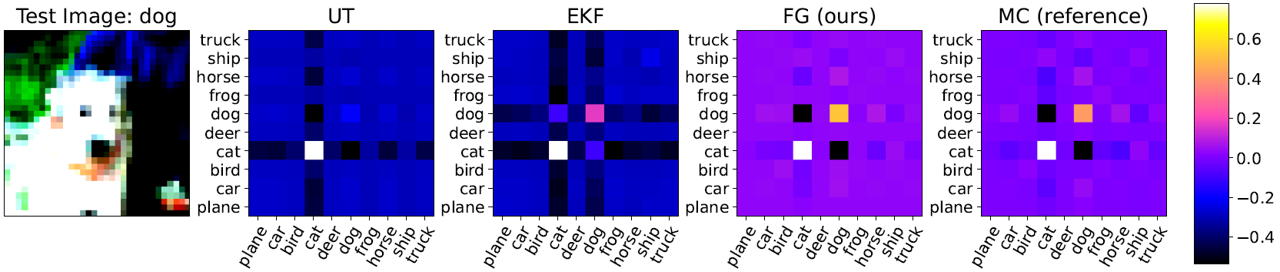}

    \includegraphics[clip,width=1.05\columnwidth]{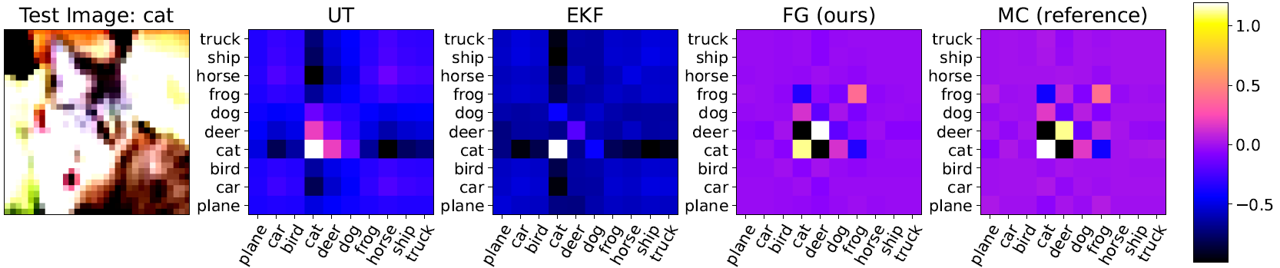}\hfill
    \includegraphics[clip,width=1.05\columnwidth]{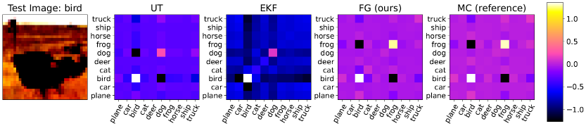}
    \caption{Examples from the CIFAR-10 experiment showing images (first and sixth column) that are corrupted with noise before input to ResNet18. Remaining columns show covariance of classifications estimated by propagating input uncertainty.}
    \label{figs:cifar}
\end{figure*}

\subsection{CIFAR-10 Classification with ResNet18}

Following our MNIST experiment, we propagated input uncertainties for a deep neural network, ResNet18, trained to classify CIFAR-10 categories. Our experimental setup follows that of the MNIST experiment. We used a trained ResNet18 with an accuracy of 93\% on the test set of CIFAR-10. Note, we excluded the LPN baseline from this experiment as it cannot propagate the uncertainties for a frozen trained network (i.e. ResNet18 would need to be retrained to additionally provide classification variances) and LPN had poor performance in our initial MNIST experiment. We evaluated the remaining methods against the Monte Carlo sampling reference using 2-Wasserstein distances for 500 example images from the CIFAR-10 test set with each of the controlled noise settings previously discussed. As in the MNIST experiment, we performed a Friedman's test followed by a Nemenyi's test to identify which experimental groups have statistically significant differences in each experimental setting. Friedman's and Nemenyi's tests were used because the 2-Wasserstein distances were not normally distributed, according to a Shapiro-Wilk tests. The results are presented in the next section.

\subsubsection{Experimental Results}

Uncertainty propagation using factor graphs provided improved performance in our CIFAR-10 experiments, as summarized quantitatively by Table~\ref{tbl:cifar} and qualitatively by Figure~\ref{figs:cifar}. Propagating input uncertainties of ResNet18 presents a challenging task due to ResNet18's high degree of non-linearity. Our factor graph approach provided statistically significant (p-value $=0.001$) improvements in performance over both baselines in all controlled noise settings except low white noise without blur (i.e. $\sigma=0.05$ and $k=1$). In the low white noise without blur, our factor graph approach only provides a statistically significant (p-value $=0.001$) improvement over the UT baseline. We presume the absence of a statistically significant improvement between our factor graph approach and the EKF baseline in the low white noise without blur setting can be attributed to the lack of input noise and high test accuracy of the trained ResNet18.

We provide Figure~\ref{figs:cifar} in addition to the quantitative CIFAR-10 results to qualitatively compare the outputs of each method. In the first two examples of Figure~\ref{figs:cifar}, the frog and ship, we see all methods perform comparably as the network is certain about the classification of the input image despite the elevated input uncertainty. Below the frog and ship, we have two examples of increasing difficulty in terms of uncertainty propagation, the plane and brown dog. In both the plane and brown dog examples, the deviation of the output covariance as estimated by the prior work compared to the reference output covariance becomes more visually prevalent in addition to the uncertainties increasing as identified by the color bar magnitudes. While the magnitude of elements within the covariance matrices estimated by the baselines deviates from the references, the set of the most uncertain classes remains the same. Finally, in the last four examples, the car, white dog, cat, and bird, the resemblance between the baseline covariances and reference covariance has severely degraded, while our factor graph approach remains difficult to distinguish from the reference across all examples.

\begin{figure}[h!]
    \centering
    \includegraphics[clip,width=0.7\columnwidth]{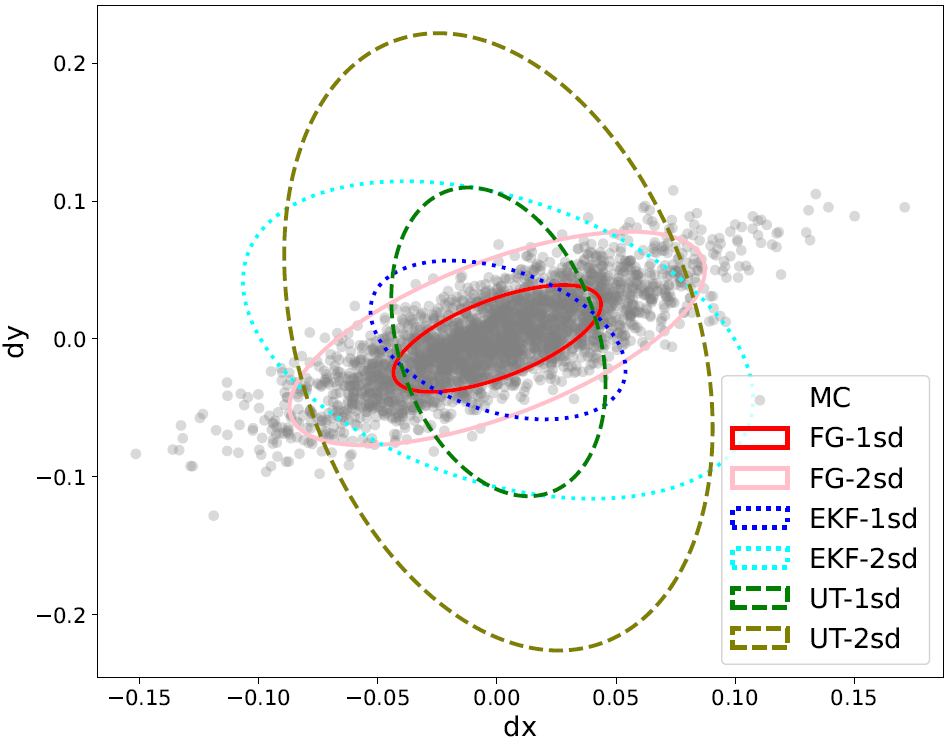}
    \includegraphics[clip,width=0.7\columnwidth]{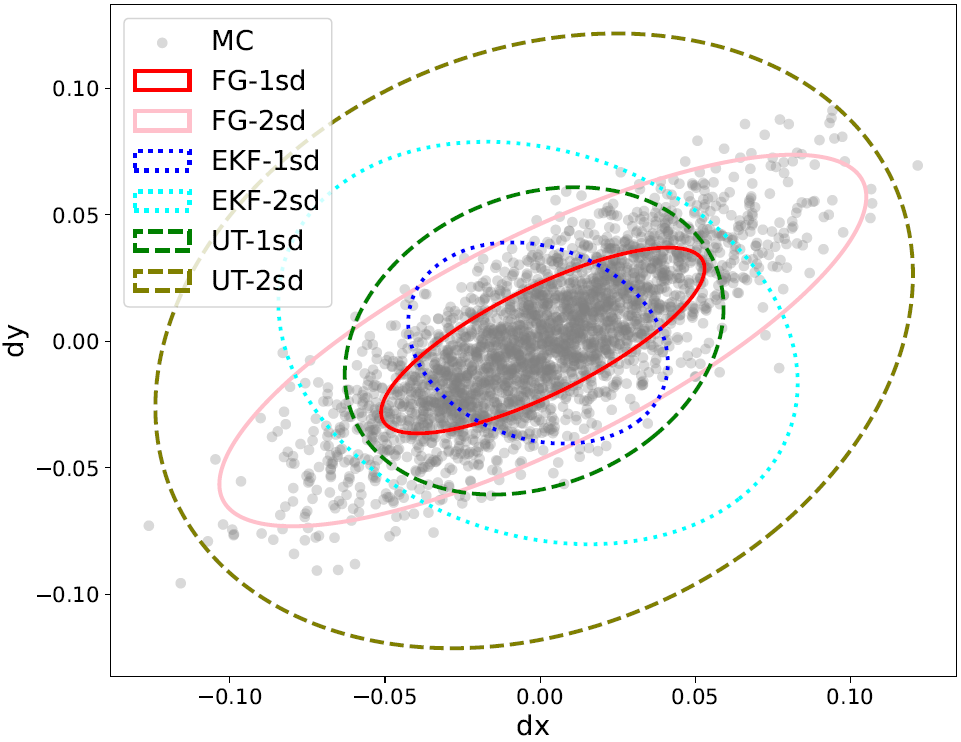}
    \includegraphics[clip,width=0.7\columnwidth]{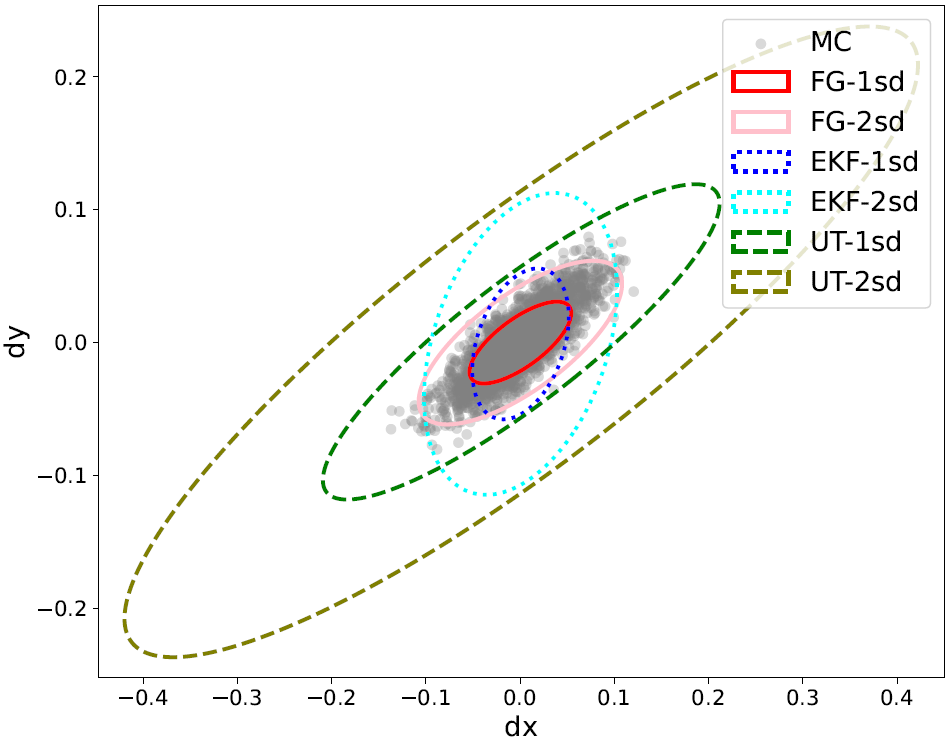}
    \caption{Three examples of uncertainty propagation on the inertial odometry network showing Monte Carlo samples as gray points with 1- and 2-standard deviation uncertainty ellipse for the output uncertainty estimated by each uncertainty propagation method.}
    \label{figs:IMU}
\end{figure}

\subsection{Inertial Odometry with ResNet18}

We propagated input uncertainties for a deep neural network, ResNet18, trained for the task of inertial odometry as our last experiment to target a more realistic use case. We followed the procedure of \cite{liu2020tlio} to train a ResNet18 for inertial odometry, wherein the network is trained to regress 2D relative displacement between two time instants given the segment of inertial measurement unit (IMU) data in between. Specifically, we revised the vanilla ResNet18 architecture into a 1D version and replaced the final fully connected layer with an MLP. As input, the inertial odometry network receives an $N$ × 6 tensor consisting of $N$ IMU measurements where each measurement contains 3 linear accelerations from the accelerometer and 3 angular velocities from the gyroscope. As output, the inertial odometry network estimates the 2D relative translation between two time instants, which is provided by the network as a two tuple, $dx$ and $dy$.

We used the publicly available M2DGR dataset \cite{yin2021m2dgr} both to train the inertial odometry network and to evaluate uncertainty propagation. M2DGR is a dataset collected by a ground robot with a full sensor-suite including the sensor we used, an Intel Realsense d435i providing 6-axis IMU data at 200 Hz. As our dataset, we used 9 trajectories from M2DGR with ground truth trajectories obtained by a Vicon Vero 2.2 motion capture system that has a localization accuracy of 1mm at 50 Hz. All but one trajectory was used to train the inertial odometry network, while the remaining trajectory was used to evaluate uncertainty propagation. In M2DGR all sensors were well-calibrated and synchronized, and their data were recorded simultaneously. As a final step we preprocess the data to be compatible with neural network training\footnote{M2DGR data processing details can be found in the Appendix}.

Note, we used the available sensor and ground truth data to estimate the input uncertainty of the IMU instead of using artificial uncertainties as in the prior ResNet18 experiment. IMU readings were assumed to be independent and identically distributed. The input uncertainty for the IMU sensor was measured as the sample covariance using the differences between IMU readings and the ground truthing system across available data. We again excluded the LPN baseline from this experiment for reasons similar to the CIFAR-10 experiment. We evaluated the remaining uncertainty propagation methods against the Monte Carlo sampling reference using 2-Wasserstein distances for 500 samples from our M2DGR test split with the measured input uncertainty of the IMU. As in the MNIST and CIFAR-10 experiments, we performed a Friedman's test followed by a Nemenyi's test to identify which experimental groups have statistically significant differences. Friedman's and Nemenyi's tests were used because the 2-Wasserstein distances were not normally distributed, according to Shapiro-Wilk tests. The results are presented in the next section.

\subsubsection{Experimental Results}

Uncertainty propagation using factor graphs provided provided statistically significant (p-value $=0.001$) improvements in performance against all baselines in our inertial odometry experiments, as summarized below and in Figure~\ref{figs:IMU}. The 2-Wasserstien distances were 0.384, 0.033, and 0.004 for UT, EKF, and FG, respectively. We again observed that our factor graph approach outperformed the prior work (i.e. UT and EKF) in this challenging and realistic experimental setting. We provide several examples in Figure~\ref{figs:IMU} in addition to 2-Wasserstien distances to qualitatively compare the outputs of each method. Each plot in Figure~\ref{figs:IMU} shows a separate test example from the inertial odometry dataset for which uncertainty propagation was evaluated. We plot 3000 Monte Carlo samples as gray points as well as the 1- and 2-standard deviation uncertainty ellipses for the output uncertainty estimated by each uncertainty propagation method. Visually, these plots demonstrate how our factor graph approach consistently provides uncertainty ellipses that better fit the Monte Carlo samples compared to the EKF or UT methods. 

Together, our experiments demonstrate the factor graph approach mitigates the risks of incorrectly propagating input uncertainties, which could result in poor decision making or complete mission failure in safety-critical applications.

\section{Conclusion}

In this paper, we developed and evaluated a novel method for input uncertainty propagation. Uncertainty propagation within modern neural networks can be challenging to model due to the complex information flows within (e.g. skip-connections). We leveraged probabilistic graphical models and factor graphs, to pose deep neural network uncertainty propagation as a non-linear optimization problem. After experimenting with multiple factor graph formulations, our reduced formulation that uses an n-ary input factor was chosen to balance between the strengths of existing baselines, analytical computations of uncertainty and input uncertainty sampling. Extensive experiments on multiple datasets demonstrate our factor graph method improves uncertainty propagation performance against all tested baselines. 

We hope this work stirs interest in using factor graphs to model deep neural network uncertainty propagation or other applications as there are many avenues for further research. Examples include, factor graph formulations for epistemic uncertainty, other models for variable nodes (e.g. Gaussian mixtures \cite{monchot2023input}), factor graph data structures that are more computationally efficient for high-dimensional variable nodes, and many others. As future work, we intend to port the approach to GPU as the current library used to perform uncertainty propagation using factor graphs is constrained to CPU, limiting the computational performance.

\section{Acknowledgements}

This research was, in part, funded by the Defense Advanced Research Projects Agency (DARPA) under Contract No. HR00112-29-0-110. The views and conclusions contained in this document are those of the authors and should not be interpreted as representing the official policies, either expressed or implied, of the U.S. Government. We would like to thank Bryan Jacobs, Jeff Stone, Sanjeev Agarwal, John Berberian, and Yeuan-Ming Sheu, for their valuable feedback.

\bibliography{aaai24}

\appendix

\newpage

\section{Inertial Odometry - M2DGR}

As our dataset, we used 9 trajectories from M2DGR with ground truth trajectories obtained by a Vicon Vero 2.2 motion capture system that has a localization accuracy of 1mm at 50 Hz. Our IMU sensor was an Intel Realsense d435i providing 6-axis IMU data at 200 Hz. Together, all trajectories within our dataset (a subset of M2DGR) comprise 2490 seconds. For network training, we use an overlapping sliding window on each available trajectory to collect input samples. As input, the inertial odometry network receives an $N × 6$ tensor consisting of $N$ IMU measurements where each measurement contains 3 linear accelerations from the accelerometer and 3 angular velocities from the gyroscope. In our ﬁnal system we choose $N = 250$ for 200 Hz IMU data. As output, the inertial odometry network estimates the 2D relative translation between two time instants $\hat{d}$, which is provided by the network as a two tuple, $d_x$ and $d_y$. We used the Mean Square Error (MSE) loss to train the inertial odometry network. The MSE loss on the training dataset is defined as:
\begin{equation} \label{eqs:training_loss}
    \mathcal{L}_{\textrm{MSE}}(d,\hat{d})=\frac{1}{n}\sum^{n}_{i=1}||d_i-\hat{d}_i||^2
\end{equation}
where $\hat{d}$ are the 2D displacement output of the network and $d$ are the ground truth displacement, $n$ is the number of data in the training set.

\end{document}